\documentclass[runningheads]{llncs}

\usepackage{eccv}

\usepackage{eccvabbrv}

\usepackage{graphicx}
\usepackage{booktabs}
\usepackage{amsmath}
\usepackage{bm}
\usepackage{algorithm, algpseudocode}

\usepackage[accsupp]{axessibility}  

\usepackage{hyperref}

\usepackage{orcidlink}
\usepackage{multirow}
\usepackage{booktabs}
\usepackage{colortbl}
\makeatletter
\def\thanks#1{\protected@xdef\@thanks{\@thanks
        \protect\footnotetext{#1}}}
\makeatother

\begin{document}

\title{CLIP-Guided Generative Networks for Transferable Targeted Adversarial Attacks}

\titlerunning{CLIP-Guided Networks for Transferable Targeted Attacks}

\author{Hao Fang\inst{1}$^{\dag}$ \and
Jiawei Kong\inst{2}$^{\dag}$ \and
Bin Chen\inst{2,3,4}$^{\#}$ \and 
Tao Dai\inst{5} \and 
Hao Wu\inst{6} \and \\
Shu-Tao Xia\inst{1,3} 
\thanks{$^{\dag}$Equal contribution.}
\thanks{$^{\#}$Corresponding author.}
}

\authorrunning{H. Fang et al.}
\institute{$^{1}$ Tsinghua Shenzhen International Graduate School, Tsinghua University \\
$^{2}$ Harbin Institute of Technology, Shenzhen\quad $^{3}$ Pengcheng Laboratory \\ $^{4}$ Guangdong Provincial Key Laboratory of Novel Security Intelligence Technologies \\ $^{5}$ Shenzhen University \quad$^{6}$ Shenzhen Digital Certificate Authority CO., Ltd \\
\email{fang-h23@mails.tsinghua.edu.cn}, \email{kongjiawei@stu.hit.edu.cn} \\
\email{chenbin2021@hit.edu.cn}, \email{daitao.edu@gmail.com}, \email{whpc79@163.com}, \email{xiast@sz.tsinghua.edu.cn}}

\maketitle
\vspace{-1.5em}
\begin{abstract}
Transferable targeted adversarial attacks aim to mislead models into outputting adversary-specified predictions in black-box scenarios. Recent studies have introduced \textit{single-target} generative attacks that train a generator for each target class to generate highly transferable perturbations, resulting in substantial computational overhead when handling multiple classes. \textit{Multi-target} attacks address this by training only one class-conditional generator for multiple classes. However, the generator simply uses class labels as conditions, failing to leverage the rich semantic information of the target class. To this end, we design a \textbf{C}LIP-guided \textbf{G}enerative \textbf{N}etwork with \textbf{C}ross-attention modules (CGNC) to enhance multi-target attacks by incorporating textual knowledge of CLIP into the generator. Extensive experiments demonstrate that CGNC yields significant improvements over previous multi-target generative attacks, e.g., a 21.46\% improvement in success rate from ResNet-152 to DenseNet-121. Moreover, we propose a masked fine-tuning mechanism to further strengthen our method in attacking a single class, which surpasses existing single-target methods. Our code is availabel at \textcolor{magenta}{\url{https://github.com/ffhibnese/CGNC_Targeted_Adversarial_Attacks}}
\end{abstract}
\section{Introduction}
\label{sec:intro}
With the rapid progress of deep learning, deep neural networks (DNNs) have been widely applied in many security-critical fields, such as autonomous driving \cite{eykholt2018robust, kong2020physgan}, financial systems \cite{sarkar2018robust}, and point cloud modeling \cite{zha2023instance, zha2024towards}. However, DNNs are corroborated to be vulnerable to adversarial attacks \cite{goodfellow2014explaining, szegedy2013intriguing, gao2024adversarial}, which attempt to fool models with adversarial examples crafted by adding imperceptible perturbations to the original inputs.
Based on the attack goal, adversarial attacks can be categorized into untargeted and targeted attacks. Untargeted attacks attempt to fool DNNs into predicting incorrect labels while targeted attacks aim at triggering the model to output the attacker-desired predictions. 

Recent investigations into the adversarial transferability \cite{andriushchenko2020square, szegedy2013intriguing, luo2021generating, athalye2018synthesizing} have demonstrated that adversarial examples crafted for a white-box surrogate model can also mislead other unseen black-box models.
Since this attack does not require access to the target model, it exposes a serious security threat to real-world applications of DNNs and motivates a wide range of studies \cite{akhtar2018threat, liu2020bias, carlini2017towards, moosavi2016deepfool}.
Despite the remarkable performance on \textit{untargeted} transferable attacks, these approaches produce unsatisfactory results for \textit{targeted} attacks due to their over-reliance on the white-box surrogate.
\begin{figure}[t]
	\centering
		\includegraphics[width=0.9\linewidth]{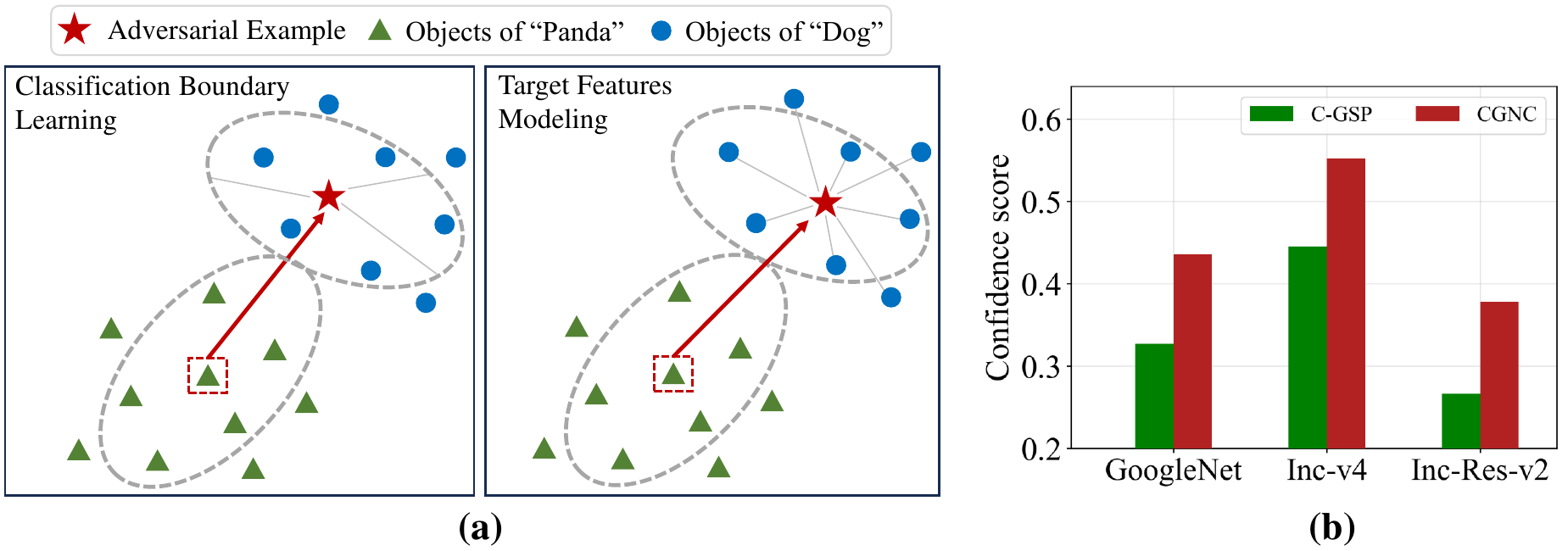}
	\centering
	\caption{(a) Targeted attacks from 'Panda' to 'Dog'. The left figure illustrates that previous multi-target methods \cite{han2019once, yang2022boosting} generate perturbations simply conditioned by class indices or one-hot vectors and only learn the classification boundary specific to the surrogate model.
 In contrast, our method exploits CLIP's meaningful guidance to effectively capture the feature distribution inherent to the target data, thereby essentially boosting the transferability.
    (b) We directly feed \textit{the scaled perturbations} generated by both our CGNC and C-GSP \cite{yang2022boosting} into three black-box classifiers. The results reveal that our generated perturbations achieve significantly higher mean confidence of the target class, demonstrating its superiority in modeling the target feature distribution.}
    \label{fig:contri}
    \vspace{-0.6cm}
\end{figure}
Existing studies on transferable targeted attacks can be categorized into \textit{instance-specific} \cite{dong2018boosting, gao2021feature, eykholt2018robust, xiong2022stochastic, wang2021enhancing, lu2020enhancing, li2020yet} and \textit{instance-agnostic} attacks \cite{xiao2018generating, luo2021generating, kong2020physgan, naseer2019cross, naseer2021generating, feng2023dynamic}. Specifically, instance-specific attacks \cite{dong2018boosting, xie2019improving, dong2019evading} iteratively perform gradient updating to craft adversarial perturbation tailored to a specific natural sample, yet producing low targeted transferability due to overfitting the white-box substitute model.
Conversely, instance-agnostic attacks learn a universal perturbation  \cite{moosavi2017universal, zhang2020understanding} or a perturbation generator \cite{poursaeed2018generative, naseer2019cross} based on data distribution rather than the specific instance, alleviating the data-specific over-fitting issues and achieving a higher adversarial transferability. 
Recent studies further explore generative attacks to produce highly transferable perturbations, which can be divided into \textit{single-target} and \textit{multi-target} attacks as follows. 

Single-target attacks \cite{naseer2019cross, naseer2021generating, feng2023dynamic, wang2023towards} exhibit impressive performance by training perturbation generators for target categories. 
However, they require training a generator for each target class, which can lead to a heavy computation burden when attacking numerous target classes. Consequently, single-target attacks are not applicable in real-world classification systems that usually contain hundreds/thousands of target classes \cite{han2019once}.  
To address this, some studies propose multi-target attacks \cite{han2019once, yang2022boosting} that train a single conditional generative model for multi-target classes. By specifying the desired label as conditioning input, the trained generator can efficiently generate the corresponding targeted adversarial perturbation. Nevertheless, these methods simply adopt class indices \cite{han2019once} or one-hot vectors \cite{yang2022boosting} of the target labels as conditions, and thus can rely solely on the classification information from the surrogate model as the guidance of the target category. Therefore, they fail to exploit the rich semantic information about the attacked category, resulting in only modest black-box fooling rates.


 In this paper, we build upon the research line of multi-target attacks by proposing a novel CLIP-guided Generative Network with Cross-attention modules (CGNC), which leverages the advanced vision-language model CLIP \cite{radford2021learning}.
 Concretely, we revisit the architecture of the conditional generators used in \cite{han2019once, yang2022boosting} and argue that the simple class conditions, \eg, class indices or handcrafted one-hot vectors, limit the transferability of the generated perturbations (see Fig. \ref{fig:contri}(a) for comparison).
 Motivated by the impressive effects of CLIP's encoded text embeddings in multi-modal learning, we introduce concise text descriptions of the target classes to encode them as class-specific representations encapsulated with abundant information, which assist the generator in learning target class distribution and ultimately lead to a fundamental transferability improvement.
 To better incorporate the text information, we improve the condition-inputting mechanism by adding the cross-attention layers that have been proven effective in models with various input modalities. 
 Results in Fig. \ref{fig:contri}(b) confirm the capability of our CGNC in capturing the target distribution.
 Besides, we propose a masked fine-tuning (MFT) technique, which fine-tunes the trained conditional generator with a fixed text condition of the desired class for further improvement in attacking a single class. 
 
 With all the above efforts, our method achieves great efficiency and scalability. When there are hundreds of target classes and single-target methods become impractical, the proposed conditional network can be utilized to achieve significant performance over previous multi-target attacks. Conversely, for scenarios involving only a few target classes, the proposed MFT mechanism enhances CGNC to outperform existing single-target methods while substantially reducing computational costs. In summary, our main contributions are as follows:
 \begin{itemize}
    \item We propose CGNC, a novel CLIP-guided generative network with cross-attention layers that fully exploits the textual knowledge provided by the advanced CLIP model for enhanced multi-target attacks. 
    \item We introduce a masked fine-tuning mechanism to improve the single-target performance by adapting the CGNC to an individual target class.
    \item Extensive experiments show that our proposed method achieves outstanding improvements in targeted transferability compared to previous state-of-the-art attack methods in a range of settings.      
\end{itemize}





\section{Related Work}

\subsection{Vision-Language Models}
Vision-language models (VLM) drawn great attention \cite{gao2024energy, gao2024inducing} due to their promising potential in learning general visual and textual representations through contrastive pre-training on large-scale image-text pairs. Given the powerful capacity of text descriptions in modeling multi-modal tasks, these models can be effectively adapted to diverse downstream tasks through appropriately formulated textual prompts, including video understanding \cite{ju2022prompting}, image manipulation \cite{chefer2022image}, and text-to-image synthesis \cite{tao2023galip}. 
Inspired by these works, we propose to harness CLIP-based multi-modal learning to 
empower our perturbation generator, facilitating more effective learning of the target feature distribution.

\subsection{Adversarial Attacks}
Among various security threats to DNNs \cite{fang2023gifd, yu2024gi, fang2024privacy}, the adversarial attack is one of the most formidable and well-known one, which can be classified into \textit{instance-specific attacks} and \textit{instance-agnostic attacks} \cite{yang2022boosting}.

\textbf{Instance-specific Attacks.}
Since the pioneering work \cite{szegedy2013intriguing} highlighted the vulnerability of neural networks, numerous gradient-based optimization methods \cite{xiong2022stochastic, goodfellow2014explaining, athalye2018synthesizing, du2019query, kurakin2016adversarial} have been proposed to craft image-dependent perturbations.
MIM \cite{dong2018boosting} integrates a momentum item into the gradient update for utilizing the previous gradient information to avoid plunging into poor local optimum. DIM \cite{xie2019improving} enhances the transferability by randomly diversifying the input pattern, and TIM \cite{dong2019evading} implements the attack by convolving the gradient with a predefined kernel. In addition, intermediate feature space \cite{inkawhich2019feature, wei2023enhancing} and classifier information \cite{inkawhich2020perturbing} are exploited to enhance attack effects, while \cite{zhao2021success} leverages logit-based loss to achieve competitive results. More advanced works \cite{che2020smgea, hang2020ensemble, he2022revisiting} consider an ensemble of multiple surrogate models to reduce over-fitting.

\textbf{Instance-agnostic Attacks.}
In contrast to instance-specific attacks, instance-agnostic attacks learn a universal perturbation \cite{moosavi2017universal, fang2024one} or a generative model \cite{poursaeed2018generative, wang2019gan, naseer2019cross, luo2021generating, naseer2021generating, yang2022boosting} for crafting adversarial examples. By modeling the global data distribution, these methods have shown better transferability and attracted more attention in recent years. Specifically, UAP \cite{moosavi2017universal} and AAA \cite{mopuri2018ask} learn a universal perturbation to fool the model based on concrete data and compressed impression, respectively. Many subsequent works, such as GAP \cite{poursaeed2018generative}, focus on boosting the transferability of adversarial attacks by training generative models. These generative attacks can be categorized into the following two types.

\textit{Single-target Attacks.}
This type of attack requires training a generator for each target class. \cite{xiao2018generating} first introduce the generative adversarial networks (GAN) \cite{goodfellow2014generative} to generate adversarial perturbations. Then, CD-AP \cite{naseer2019cross} concentrates on the domain-invariant adversaries and launches highly transferable cross-domain attacks using a relativistic supervisory. TTP \cite{naseer2021generating} modifies the loss function and proposes to match the target distribution to mitigate over-fitting to the surrogate model. Subsequent methods achieve better transferability based on a dynamic network with pattern injection \cite{feng2023dynamic} or a feature discriminator \cite{wang2023towards}.

\textit{Multi-target Attacks.}
MAN \cite{han2019once} notes that when dealing with numerous classes, single-target attacks inevitably suffer from extreme computational burdens as they need to train the same number of models as multiple target classes. Therefore, single-target attacks become impractical in attacking real classification systems that often have hundreds of categories. To pursue the extreme speed and storage, MAN trains only \textit{one} model for 1000 target categories from ImageNet \cite{deng2009imagenet}. However, the excess of the target class severely degrades the targeted transferability. 
C-GSP \cite{yang2022boosting} improves the performance
by designing a hierarchical partition mechanism to divide all classes into a feasible number of subsets and train generators for each subset. Nonetheless, C-GSP simply conditions the generator with one-hot vectors and only utilizes the classification information to train the generator, failing to use the semantic knowledge of the target class. To overcome the limitation, we propose to incorporate the text information provided by CLIP as significant guidance for the target class.

\section{Method}
In this section, we first introduce the preliminaries of targeted transferable attacks and present the basic paradigm of generative attack methods. Then, we elaborate on the proposed CGNC, which remarkably enhances multi-target black-box attacks. Finally, we detail the proposed masked fine-tuning technique that strengthens our method in single-class attacks.

\subsection{Preliminary}
We denote the white-box image classifier parameterized with $\theta$ as $f_{\theta}: \mathcal{X} \rightarrow \mathcal{Y}$, where $\mathcal{X} \subset \mathbb{R}^{N \times H\times W}$ represents the image domain and $\mathcal{Y}
\subset \mathbb{R}^{L}$ is the output confidence score of different classes ($H, W, N, L$ being height, width, number of channels, and class number). 
Given a natural image $\bm{x} \in \mathcal{X}$ and the attacker's desired label $c_t \in \mathcal{C}$, the transferable targeted adversarial attacks attempt to craft the imperceptible perturbation $\bm{\delta}$ based on the accessible surrogate model $f_{\theta}$ to mislead another unseen victim model $F_{\phi}$ into predicting $c_t$, \ie, $\arg\max_{i \in \mathcal{C}}F_{\phi}(\bm{x} + \bm{\delta})_{i} = c_t$.
Concurrently, the $l_{\infty}$ norm is employed to ensure the adversarial samples are indistinguishable from the original images by constraining the perturbation within the range $\epsilon$ by $\Vert \bm{\delta} \Vert_{\infty} \leq \epsilon$.

To boost the fooling rate of targeted black-box attacks, single-target attacks \cite{poursaeed2018generative, naseer2019cross, naseer2021generating} utilize powerful generative models to learn the target data distribution and achieve higher transferability.
However, these methods consume great computation time and resources for multi-target scenarios, making them impractical for real-world attacks. C-GSP \cite{yang2022boosting} effectively solves this issue by formulating the multi-target attacks as learning a class-conditional generator $G_w$ with weights $w$, which is capable of generating perturbations for any target class. Given an unlabeled training dataset $\mathcal{X}_s$, the optimization objective is as follows:

\begin{equation}
\begin{split}
    \min\limits_{w} &\mathbb{E}_{\bm{x}_{s}\sim \mathcal{X}_{s}, c\sim \mathcal{C}}[{\mathcal{L}\big(f_{\theta}(\bm{x}_{s} + G_{w}(\bm{x}_{s}, c)\big), c)}], \;
  \text{s.t. } \|G_{w}(\bm{x}_{s}, c) \|_{\infty} \leq \epsilon,
\end{split}
\label{eq:target}
\end{equation}

\noindent where $\mathcal{L}(\cdot, \cdot)$ denotes the cross-entropy (CE) loss. By minimizing the loss of specified classes using various unlabeled images from $\mathcal{X}_{s}$, we optimize the parameters $w$ of the generative model and finally obtain a targeted conditional generator that can generate perturbation for any given clean image $\bm{x}_{t}$ from the test dataset $\mathcal{X}_t$. Specifically, the adversary can simply specify a target label $c$ and craft an corresponding adversarial example via $\bm{x}_{adv} = \bm{x}_{t} + G_{w}(\bm{x}_{t}, c)$.

However, current multi-target methods \cite{han2019once, yang2022boosting} simply condition the generator with class labels and learn the target distribution only relying on the classification information of the surrogate model, thus not fully exploiting the specific characteristics of the target category. Inspired by the efficiency of vision-language (VL) learning \cite{radford2021learning ,tao2023galip}, we propose a novel generative network that leverages sufficient prior knowledge from the powerful CLIP model by incorporating textual-modality information to promote the target class modeling, which greatly enhances the multi-target transferable attacks. 


\begin{figure*}[t]
\begin{center}
\includegraphics[width=1\linewidth]{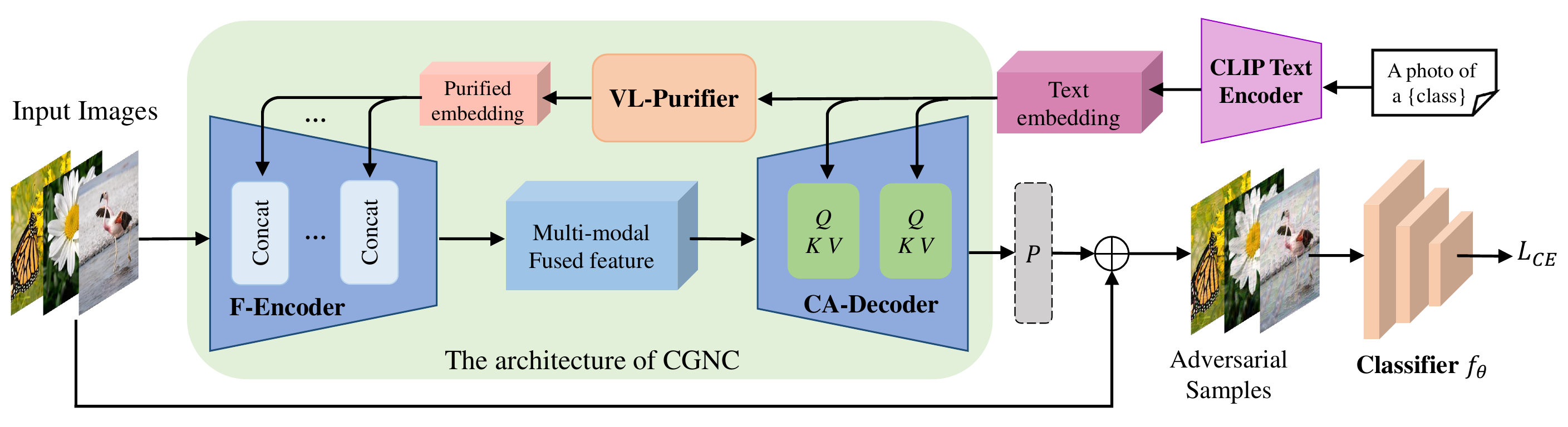}
\end{center}
\caption{An overview of our proposed architecture of CGNC. Equipped with the three exquisite modules VL-Purifier, F-Encoder, and CA-Decoder, the generator fully leverages the textual representations encoded by CLIP as auxiliary information about the target classes to better probe their data distribution and achieve superior attack effects.}
\label{fig:pipeline}
\end{figure*}

\subsection{CLIP-Guided Generative Network}
The proposed generative model architecture is presented in Fig. \ref{fig:pipeline}. Specifically, CGNC is composed of a Vision-Language feature Purifier (VL-Purifier), a feature Fusion Encoder (F-Encoder), and a Cross-Attention based Decoder (CA-Decoder). We also provide the pseudocode of the training procedure in Algorithm \ref{alg1}. Next, we illustrate the design of each module as follows.

\textbf{Vision-language feature purifier (VL-Purifier)}. To utilize CLIP to produce semantic embeddings, we first feed in CLIP's text encoder $\Phi$ with queries $\bm t_c$ that follow the handcrafted template "a photo of a \{class\}", which has shown effectiveness in many tasks \cite{radford2021learning, huang2023sentence}. Since the obtained embeddings $\bm{e}_{t} \in \mathbb{R}^{B\times 512}$ ($B$ being the batch size) in CLIP's vision-language space are generic representations of the target classes and are not yet tailored to our learning task, we refine them using the VL-Purifier, which is composed of several blocks consisting of a fully-connected layer and a spectral normalization layer. Through this module, we translate the encoded embeddings into more meaningful representations $\bm{e}_{t}^{*} \in \mathbb{R}^{B\times 16}$, thereby facilitating the subsequent step of feature fusion.

\textbf{Feature fusion encoder (F-Encoder)}. This module aims to fuse the purified features $\bm{e}_{t}^{*}$ with the learned visual representations. Firstly, a batch of input images $\bm{x}_{s}$ is encoded to capture the visual concepts $\bm{h}_{s} \in \mathbb{R}^{B \times N'\times H'\times W'}$. Then, we expand the text embedding $\bm{e}_{t}^{*} \in \mathbb{R}^{B\times 16}$ into $\bm{e}_{t}^{*}{'} \in \mathbb{R}^{{B\times 16\times H'\times W'}}$, which are then integrated with the extracted visual concepts $\bm{h}_{s}$ through channel-wise concatenation to obtain the fused representations, \ie, $\bm{m} \in \mathbb{R}^{B\times (N'+16) \times H'\times W'}$. 
Subsequently, $\bm{m}$ undergoes further downsampling, and the resultant features are again concatenated with the expanded embedding $\bm{e}_{t}^{*}{'}$. By repeating this operation several times, we effectively fuse the visual concepts of input images and the purified CLIP's embedding of the target classes. This mechanism fully exploits both the instance-level and class-level information from visual and textual modalities, thus encouraging the generation of perturbations with better semantic patterns and higher transferability.

\begin{algorithm}[t]
    \caption{Pseudocode of Training the CLIP-guided Generative Network}\label{alg1}
\begin{algorithmic}[1]
\Require 
    $\mathcal{X}_s$: the training data; 
    $\mathcal{C}$: the target label space;     
    $\mathcal{T}$: the text prompts set;
    $f_{\theta}$: the surrogate model; 
    $\Phi$: the CLIP's text encoder;
    $N$: the max iteration;
\Ensure the perturbation generator $G_{w}$;
\For{$i \leftarrow 0$ to $N$}
\State Sample a batch of images $\bm{x}_s\sim \mathcal{X}_s$;
\State Obtain $\bm{x}_s'$ by processing $\bm{x}_s$ with data augmentation;
\State Sample a batch of target labels $c \sim \mathcal{C}$;
\State Obtain the corresponding text prompts $\bm{t}_c$ from $\mathcal{T}$;
\State Compute the text embedding $\bm{e}_{t}$ by feeding $\bm{t}_c$ into $\Phi$;
\State Obtain perturbed images by $\bm{x}_{adv} = \bm{x}_s+G_w(\bm{x}_s,\bm{e}_t)$, $\bm{x'}_{adv} = \bm{x'}_s+G_w(\bm{x'}_s,\bm{e}_t)$;
\State Forward pass $\bm{x}_{adv}$, $\bm{x'}_{adv}$ to $f_{\theta}$ and compute the loss in Eq.~\eqref{eq:objective};
\State  Perform gradient backpropagation and update the generator $G_{w}$;
\EndFor
\State \textbf{return} the trained generator $G_{w}$

\end{algorithmic}
\end{algorithm}

\textbf{Cross-Attention based Decoder(CA-Decoder)}.
Given the multi-modal fused features from the previous module, the decoder attempts to translate them into perturbations of the target class. The network backbone is realized based on the decoder used in previous works \cite{naseer2021generating, yang2022boosting}. To fully explore the semantic priors brought by the CLIP model, we enhance the underlying backbone by introducing the cross-attention mechanism, which is proven to be effective for many multi-modal learning tasks. Specifically, we incorporate the textual embedding $\bm{e}_t$ from CLIP's latent space into our decoder via the cross-attention layer:


\begin{equation}
\begin{split}
Q=\bm z_{t}W_{q}, K=&\ \bm e_{t}W_{k}, V = \bm e_{t}W_{v}, \\
Attention(Q,K,V)&=softmax(\frac{QK^{T}}{\sqrt{d}}) \cdot V,  
\end{split}
\label{eq:cross_attn}
\end{equation}

\noindent where $\bm{z}_t \in \mathbb{R}^{B\times d_{\alpha}}$ denotes the flattened intermediate features of the decoder, $W_{q}\in \mathbb{R}^{d_{\alpha}\times d}$, $W_{k}\in \mathbb{R}^{512\times d}$, $W_{v}\in \mathbb{R}^{512\times d}$ are learnable parameters. 
Similar to C-GSP, we post-process the output $\bm o$ via the $\mathrm{tanh}(\cdot)$ smooth projection to obtain the $\ell_{\infty}$ constrained perturbation with budge $\epsilon$, \ie, $\bm \delta = P(\bm o) = \epsilon \cdot \mathrm{tanh}(\bm o)$.


\textbf{Optimization objective.} Based on the proposed CLIP-guided architecture,  the optimization objective can be formulated as:
\begin{equation}
    w^{*} \leftarrow \arg\min_{w}{\mathcal{L} \Big(f_{\theta}\big(\bm{x}_{s} + G_{w}(\bm{x}_{s}, \Phi(\bm t_c))\big), c\Big)},
\label{eq:objective}
\end{equation}
where $\bm t_c$ represents the corresponding text prompts of the target classes $c$, $\mathcal{L}(\cdot, \cdot)$ denotes the cross-entropy loss. By encouraging the network to craft adversarial samples that can misguide the surrogate model's output toward the desired labels, the generator learns the data distribution of the target class and thus exhibits great generalizability in producing perturbation for any input data.

\begin{figure}[t]
	\centering
	\begin{subfigure}{0.55\linewidth}
		\centering
		\includegraphics[width=\linewidth]{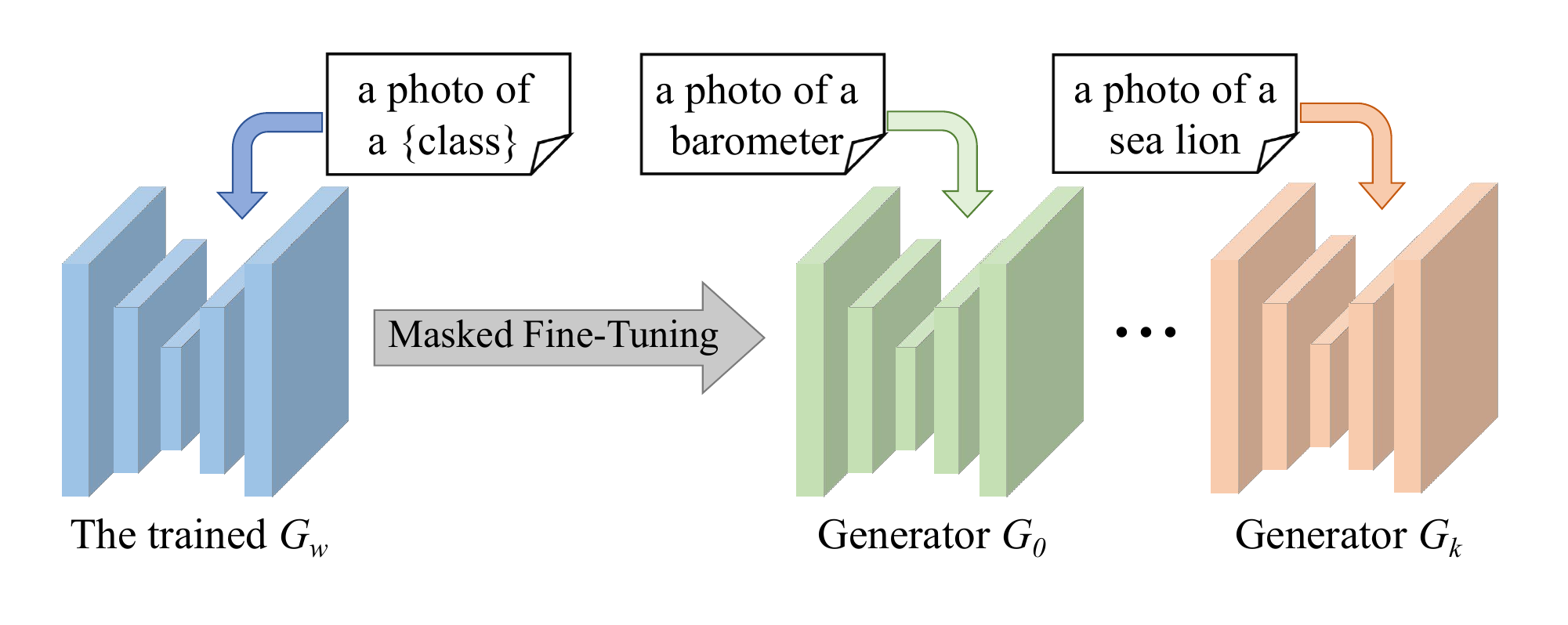}
		\caption{}
		\label{mft}
	\end{subfigure}
	\centering
	\begin{subfigure}{0.44\linewidth}
		\centering
		\includegraphics[width=\linewidth]{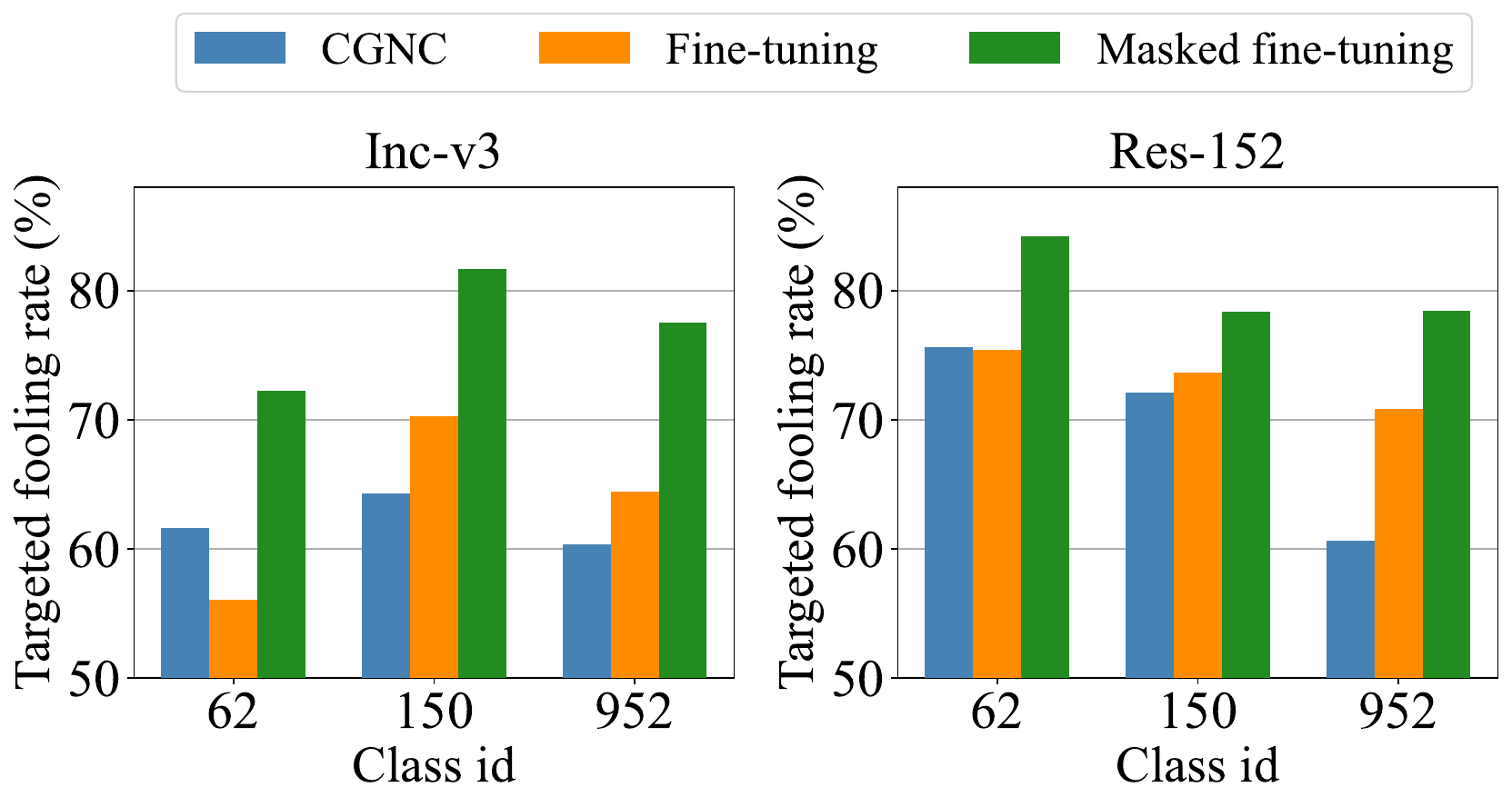}
		\caption{}
		\label{mft:results}
	\end{subfigure}
	\centering
	\caption{Illustration of the proposed masked fine-tuning mechanism. (a) We fix the condition input with different text prompts to fine-tune the trained conditional generator $G_{w}$ into multiple generators for single-target attacks. (b) The fooling rate of several target classes with Inc-v3 and Res-152 as substitute models respectively. The results indicate that direct fine-tuning yields inadequate results for certain classes due to overfitting. We efficiently resolve this issue via a patch-wise random mask operation. 
    }
    \vspace{-0.4cm}
\end{figure}

\subsection{Masked Fine-Tuning Mechanism}
In addition to the proposed CLIP-guided generator for multi-target attack scenarios, we also design a \textit{single-target} variant for further improved performance. As depicted in Fig. \ref{mft}, we fix the conditioning input with the text description of a specific target class, and fine-tune the trained multi-target generator by resolving the objective function in Eq. (\ref{eq:objective}) on the unlabeled training data $\mathcal{X}_{s}$. This strategy further enhances the attack effects by fine-tuning the conditional generator to specialize in a specific target class. 

However, we encounter the overfitting problem that leads to limited improvement in success rates or even performance degradation for certain target classes. This problem is partly attributed to the fact that the generated adversarial perturbation sometimes heavily focuses on specific regions of the input image \cite{wang2023towards}. To alleviate this issue, we adopt a patch-wise random mask operation to post-process the adversarial perturbation produced by the generator, which brings a notable increase in the targeted fooling rate as shown in Fig. \ref{mft:results}. This technique improves the capability and flexibility of our method, as an adversary can perform fine-tuning to further augment the generator for single-target attacks, achieving a better trade-off between efficiency and performance compared to previous single-target methods (See Section \ref{sec:single_target} for detailed analysis).

\section{Experiments}
\label{sec:blind}

To validate the effectiveness of our proposed CGNC in boosting transferability, we conduct extensive experiments on various black-box models in a range of scenarios. Please see the Appendix for more experimental results.

\subsection{Experimental Settings}

\textbf{Datasets.}
Following \cite{yang2022boosting, feng2023dynamic}, we train the generator on the ImageNet training set \cite{deng2009imagenet} and evaluate the attack performance using ImageNet-NeurIPS (1k) dataset proposed by \cite{Nips2017}. We also conduct experiments in a more realistic cross-domain scenario where we train the generator and generate perturbations on MS-COCO \cite{lin2014microsoft} or Comics \cite{comics} while evaluating on target classifier trained on ImageNet.

\textbf{Victim Models.}
We use various victim models with different architectures. Specifically, the naturally trained models include Inception-v3 (Inc-v3) \cite{szegedy2016rethinking}, Inception-v4 (Inc-v4) \cite{szegedy2017inception}, Inception-ResNet-v2 (Inc-Res-v2) \cite{szegedy2017inception}, ResNet-152 (Res-152) \cite{he2016identity}, DenseNet-121 (DN-121) \cite{huang2017densely}, GoogleNet \cite{szegedy2015going}, and VGG-16 \cite{simonyan2014very}. 

For further evaluation, we also analyze on the robust-trained models, including adv-Inception-v3 ($\textrm{Inc-v3}_\textrm{adv}$) \cite{goodfellow2014explaining}, ens-adv-Inception-ResNet-v2 ($\textrm{IR-v2}_\textrm{ens}$) \cite{hang2020ensemble}, and robust-trained ResNet-50 \cite{geirhos2018imagenet, hendrycks2019augmix}, dubbed as $\textrm{Res50}_\textrm{SIN}$ (trained on stylized ImageNet), $\textrm{Res50}_\textrm{IN}$ (trained on the mixture of stylized and Nature ImageNet), $\textrm{Res50}_\textrm{fine}$ ($\textrm{Res50}_\textrm{IN}$ plus further finetuning  with an auxiliary dataset \cite{geirhos2018imagenet}), and $\textrm{Res50}_\textrm{Aug}$ (trained with the advanced data augmentation Augmix \cite{hendrycks2019augmix}). 


\textbf{Baseline Attacks.}
We reveal the superiority of the proposed CGNC in enhancing multi-target attacks by comparing our method with multiple competitive baselines, including MIM \cite{dong2018boosting}, DIM \cite{xie2019improving}, SIM \cite{lin2019nesterov}, DIM \cite{dong2019evading}, Logit \cite{zhao2021success}, SU \cite{wei2023enhancing}, and the state-of-the-art (SOTA) multi-target generative attacks C-GSP \cite{yang2022boosting} in Section \ref{sec:multi_attack}. For SU attack \cite{wei2023enhancing}, we choose to compare with its best version DTMI-Logit-SU. Besides, we also provide a comparison of the proposed single-target variant of CGNC with the existing single-target attack methods, \ie, GAP \cite{poursaeed2018generative}, CD-AP \cite{naseer2019cross}, TTP \cite{naseer2021generating}, and DGTA-PI\cite{feng2023dynamic} in Section \ref{sec:single_target}.

\textbf{Implementation Details.}
Following previous works \cite{yang2022boosting, feng2023dynamic}, we employ Inc-v3 and Res-152 as surrogate models to guide the generation of adversarial noises. Unless stated otherwise, the perturbation budget $\epsilon$ is 16/255. We conduct 10 epochs of training for the generator using a learning rate of 2e-4. During the masked fine-tuning process, we maintain the same learning rate and apply a mask ratio of 0.2 to fine-tune the text-conditional generator for an additional 5 epochs, using different text prompts to obtain multiple corresponding generators.

\subsection{Multi-Target Transferability Evaluation}
\label{sec:multi_attack}
To align the experimental setup to former works \cite{yang2022boosting, feng2023dynamic}, we first target 8 different classes used in \cite{zhang2020understanding} to reach the multi-target black-box attack testing protocol. The average attack success rates (ASR) of the 8 target categories are presented as evaluation metrics. Attack performance under larger numbers of target categories are presented in the next section.

\textbf{Attacks against regularly trained models.}
We first perform attacks on normal models to evaluate the multi-target attack performance. By observing the results in Table \ref{tab:main}, we demonstrate that the proposed CGNC consistently achieves significant improvement compared with previous methods. Specifically, our method achieves an average improvement of 17.88\% and 10.08\% in ASR over the C-GSP \cite{yang2022boosting} attack regarding Res-152 and Inc-v3 as surrogate models respectively, demonstrating the superiority of leveraging the rich prior knowledge from CLIP's text embedding.
We also note that the iterative methods obtain nearly 100\% while-box ASR while receiving poor performance on black-box models due to overfitting the classification boundaries of surrogate models.

\begin{table}[t]

\setlength{\tabcolsep}{8pt}
  \centering
  \caption{Attack success rates (\%) for multi-target attacks against regularly trained models on ImageNet NeurIPS validation set. * represents white-box attacks.}
      \vspace{-0.5em}
    \resizebox{0.97\linewidth}{!}{\begin{tabular}{c|c|c|c|c|c|c|c|c}
    \toprule 
    Source   & Method   & Inc-v3   & Inc-v4   & Inc-Res-v2    & Res-152  & DN-121   & GoogleNet       & VGG-16 \\ [0.05pt]
    \hline
    \multirow{10}[0]{*}{Inc-v3} & MIM      & \textbf{99.90}$^*$ & 0.80     & 1.00     & 0.40     & 0.20     & 0.20     & 0.30 \\
             & TI-MIM   & 98.50$^*$    & 0.50     & 0.50     & 0.30     & 0.20     & 0.40     & 0.40 \\
             & SI-MIM   & 99.80$^*$    & 1.50     & 2.00     & 0.80     & 0.70     & 0.70     & 0.50 \\
             & DIM      & 95.60$^*$    & 2.70     & 0.50     & 0.80     & 1.10     & 0.40     & 0.80 \\
             & TI-DIM   & 96.00$^*$    & 1.10     & 1.20     & 0.50     & 0.50     & 0.50     & 0.80 \\
             & SI-DIM   & 90.20$^*$    & 3.80     & 4.40     & 2.00     & 2.20     & 1.70     & 1.40 \\
             & Logit    & 99.60$^*$    & 5.60     & 6.50     & 1.70     & 3.00     & 0.80     & 1.50 \\
             & SU       & 99.59$^*$    & 5.80     & 7.00     & 3.35     & 3.50     & 2.00     & 3.94 \\
             & C-GSP    & 93.40$^*$    & 46.58    & 36.74    & 41.60    & 46.40    & 40.00    & 45.00 \\
             & \cellcolor{black!10}CGNC& \cellcolor{black!10}96.03$^*$& \cellcolor{black!10}\textbf{59.43}&\cellcolor{black!10}\textbf{48.06}&\cellcolor{black!10}\textbf{42.48}&\cellcolor{black!10}\textbf{62.98}&\cellcolor{black!10}\textbf{51.33}&\cellcolor{black!10}\textbf{52.54}\\
             \hline
    \multirow{10}[0]{*}{Res-152} & MIM      & 0.50     & 0.40     & 0.60     & \textbf{99.70}$^*$ & 0.30     & 0.30     & 0.20 \\
             & TI-MIM   & 0.30     & 0.30     & 0.30     & 96.50$^*$    & 0.30     & 0.40     & 0.30 \\
             & SI-MIM   & 1.30     & 1.20     & 1.60     & 99.50$^*$    & 1.00     & 1.40     & 0.70 \\
             & DIM      & 2.30     & 2.20     & 3.00     & 92.30$^*$    & 0.20     & 0.80     & 0.70 \\
             & TI-DIM   & 0.80     & 0.70     & 1.00     & 90.60$^*$    & 0.60     & 0.80     & 0.50 \\
             & SI-DIM   & 4.20     & 4.80     & 5.40     & 90.50$^*$    & 4.20     & 3.60     & 2.00 \\
             & Logit    & 10.10    & 10.70    & 12.80    & 95.70$^*$    & 12.70    & 3.70     & 9.20 \\
             & SU       & 12.36 & 11.31 & 16.16 & 95.08$^*$ & 16.13 & 6.55 & 14.28 \\
             & C-GSP    & 37.70    & 33.33    & 20.28    & 93.20$^*$    & 64.20    & 41.70    & 45.90 \\
             & \cellcolor{black!10}CGNC&\cellcolor{black!10}\textbf{53.39}&\cellcolor{black!10}\textbf{51.53}&\cellcolor{black!10}\textbf{34.24}&\cellcolor{black!10}95.85$^*$&\cellcolor{black!10}\textbf{85.66}&\cellcolor{black!10}\textbf{62.23}&\cellcolor{black!10}\textbf{63.36}\\
             \bottomrule
    \end{tabular}%
    }
    \vspace{-1.8em}
  \label{tab:main}%
\end{table}%

\begin{table}[t]
\setlength{\tabcolsep}{8pt}
  \centering
  \caption{Comparison of the proposed CGNC with the SOTA multi-target attacks against models with robust training mechanism on ImageNet NeurIPS validation set. }
    \resizebox{0.86\linewidth}{!}{\begin{tabular}{c|c|c|c|c|c|c|c}
    \toprule
    Source   & Method   & $\textrm{Inc-v3}_\textrm{adv}$ & $\textrm{IR-v2}_\textrm{ens}$ & $\textrm{Res50}_\textrm{SIN}$ & $\textrm{Res50}_\textrm{IN}$ & $\textrm{Res50}_\textrm{fine}$ & $\textrm{Res50}_\textrm{Aug}$ \\ [0.05pt]
    \hline
    \multirow{10}[0]{*}{Inc-v3} & MIM      & 0.16     & 0.10     & 0.20     & 0.27     & 0.44     & 0.19 \\
             & TI-MIM   & 0.21     & 0.19     & 0.33     & 0.49     & 0.68     & 0.31 \\
             & SI-MIM   & 0.13     & 0.19     & 0.26     & 0.43     & 0.63     & 0.29 \\
             & DIM      & 0.11     & 0.09     & 0.16     & 0.33     & 0.39     & 0.19 \\
             & TI-DIM   & 0.15     & 0.13     & 0.16     & 0.21     & 0.33     & 0.14 \\
             & SI-DIM   & 0.19     & 0.21     & 0.43     & 0.71     & 0.84     & 0.46 \\
             & Logit    & 0.30     & 0.30     & 0.70     & 1.23     & 3.14     & 0.86 \\
             & SU       & 0.49     & 0.41     & 0.84     & 1.75     & 3.55     & 1.04 \\
             & C-GSP    & 20.41    & 18.04    & 6.96     & 33.76    & 44.56    & 21.95 \\
             & \cellcolor{black!10}CGNC & \cellcolor{black!10}\textbf{24.36} & \cellcolor{black!10}\textbf{22.54} & \cellcolor{black!10}\textbf{8.85} & \cellcolor{black!10}\textbf{40.83} & \cellcolor{black!10}\textbf{52.18} & \cellcolor{black!10}\textbf{22.85} \\
             \hline
    \multirow{10}[0]{*}{Res-152} & MIM      & 0.19     & 0.15     & 0.28     & 1.58     & 2.75     & 0.78 \\
             & TI-MIM   & 0.61     & 0.73     & 0.50     & 2.51     & 4.75     & 1.76 \\
             & SI-MIM   & 0.24     & 0.24     & 0.39     & 0.66     & 0.84     & 0.36 \\
             & DIM      & 0.63     & 0.37     & 0.94     & 8.50     & 14.22    & 3.77 \\
             & TI-DIM   & 0.23     & 0.30     & 0.28     & 0.76     & 1.49     & 0.49 \\
             & SI-DIM   & 0.71     & 0.71     & 0.75     & 2.73     & 3.89     & 1.37 \\
             & Logit    & 1.15     & 1.18     & 1.65     & 6.70     & 15.46    & 5.93 \\
             & SU       & 2.12 & 1.20 & 1.95 & 7.53 & 21.14 & 6.95 \\
             & C-GSP    & 14.60    & 16.01    & 16.84    & 60.30    & 65.51    & 42.88 \\
             & \cellcolor{black!10}CGNC & \cellcolor{black!10}\textbf{22.21} & \cellcolor{black!10}\textbf{26.71} & \cellcolor{black!10}\textbf{29.83} & \cellcolor{black!10}\textbf{79.80} & \cellcolor{black!10}\textbf{84.05} & \cellcolor{black!10}\textbf{63.75} \\
             \bottomrule
    \end{tabular}%
    }
  \label{tab:adv}%
\end{table}%

\textbf{Attacks under defense strategies.}
For a more thorough analysis and comparison, we then compare these attacks under several widely used defenses. Firstly, we consider attacking six robustly trained networks and the experiment results are in Table \ref{tab:adv}. Generally, our method is still able to deceive the black-box classifiers into predicting the specified classes when dealing with robustness-augmented models, significantly outperforming former multi-target attacks, \eg, a 20.87 \% increase of fooling rate from Res-152 to $\textrm{Res50}_\textrm{Aug}$.

Next, we evaluate the performance of different approaches on models with input preprocessing-based defenses, including a set of image smoothing mechanisms \cite{ding2019advertorch} and JPEG compression \cite{dziugaite2016study} algorithms. As shown in Table \ref{tab:multi_defense}, although these defenses eliminate certain valid information in the adversarial samples, the CLIP-empowered CGNC still demonstrates excellent capability in boosting targeted transferability. Particularly when the substitute model is Res-152, CGNC achieves an average fooling rate of 71.18\% and 80.34\% on smoothing methods and JPEG compression, while C-GSP only reaches 49.05\% and 56.16\% respectively, verifying the stability and robustness of the proposed method.

\begin{figure}[t]
	\centering
	\includegraphics[width=\linewidth]{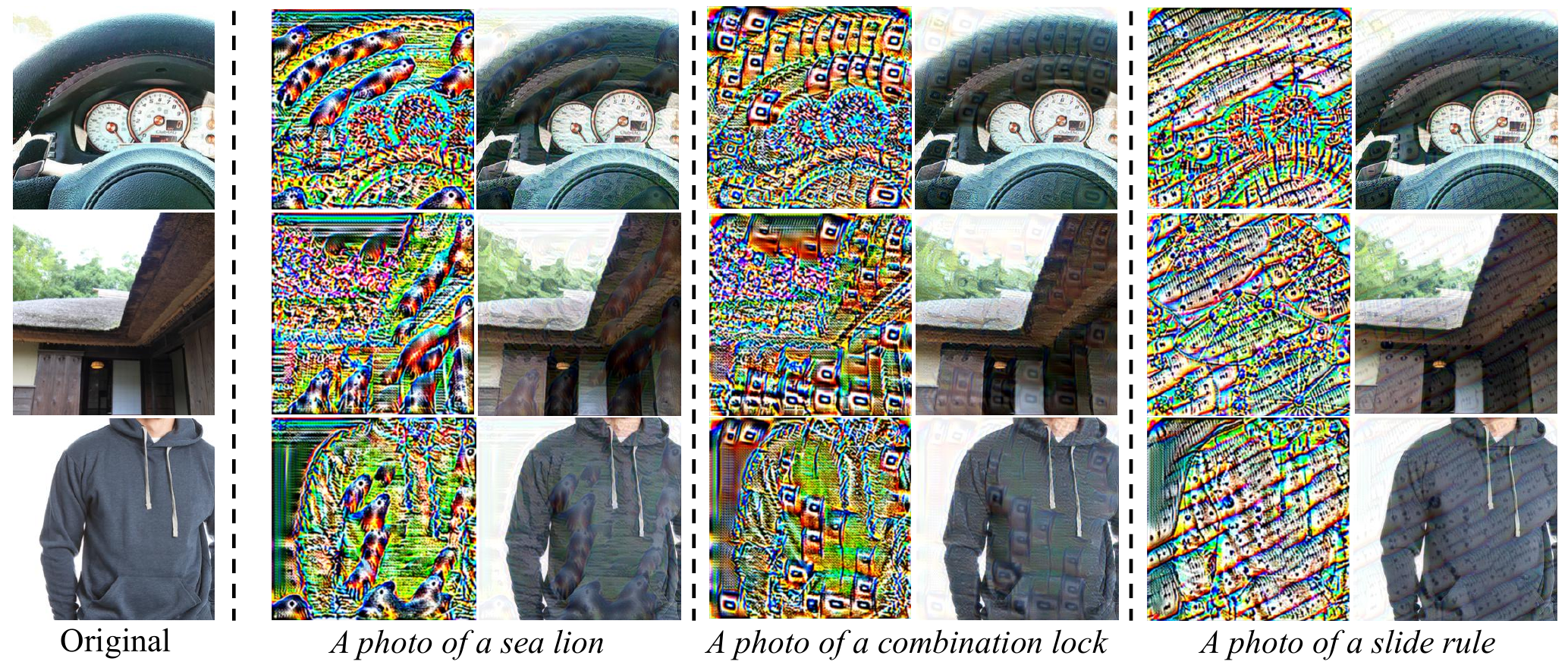}
	\caption{Visualization results of different input images for different targets. For each text prompt of the target class, the left column shows the perturbation generated by our CGNC while the right column displays the corresponding adversarial examples. }
	\label{fig:visual}

\end{figure}

\textbf{Perturbation Visualization.}
We present visualization results in Fig. \ref{fig:visual} to unveil the principle of our method. Upon careful inspection, we can see that the generated perturbations mainly focus on the semantic regions of the input images and contain sufficient semantic patterns specific to the target category. For instance, when the condition is \textit{a photo of a sea lion}, the resulting perturbations indeed carry rich patterns closely resembling this marine animal. We also observe that the pattern changes in accordance with the text prompts, which validates our conditioning mechanism of the CLIP-encoded textual embedding. 

\begin{table}[t]
\setlength{\tabcolsep}{6pt}
  \centering
  \caption{Comparison of our method with C-GSP under different defense strategies. Q indicates the quality factor in JPEG compression. Here the target model is DN-121 and results for more victim models are in the Appendix.}
  \resizebox{0.92\linewidth}{!}{
    \begin{tabular}{cccccccccc}
    \toprule
    \multirow{2}[0]{*}{Source} & \multirow{2}[0]{*}{Method} & \multicolumn{3}{c}{Smoothing}  & \multicolumn{5}{c}{JPEG compression} \\
    \cmidrule(lr){3-5} \cmidrule(l){6-10}
             &          & Gaussian & Medium   & Average  & Q=70 & Q=75 & Q=80 & Q=85 & Q=90 \\
             \hline
             \noalign{\vskip 1.8pt}
    \multirow{2}[0]{*}{Inc-v3} & C-GSP    & 36.59    & 46.91    & 39.86    & 49.41    & 50.55    & 51.95    & 52.84    & 53.39 \\
             & CGNC     & \textbf{43.03} & \textbf{55.35} & \textbf{45.94} & \textbf{58.35} & \textbf{59.25} & \textbf{60.28} & \textbf{61.29} & \textbf{61.94} \\
             \hline
             \noalign{\vskip 1.8pt}
    \multirow{2}[0]{*}{Res-152} & C-GSP    & 43.21    & 56.38    & 47.55    & 53.03    & 54.54    & 56.13    & 57.75    & 59.36 \\
             & CGNC     & \textbf{64.44} & \textbf{79.84} & \textbf{69.25} & \textbf{77.38} & \textbf{78.81} & \textbf{80.54} & \textbf{82.05} & \textbf{82.94} \\
             \bottomrule
    \end{tabular}%
    }
  \label{tab:multi_defense}%
  \vspace{-0.5em}
\end{table}%

\begin{figure*}[t]
\begin{minipage}[t]{0.48\linewidth}
  \centering
  \includegraphics[width=0.85\textwidth]{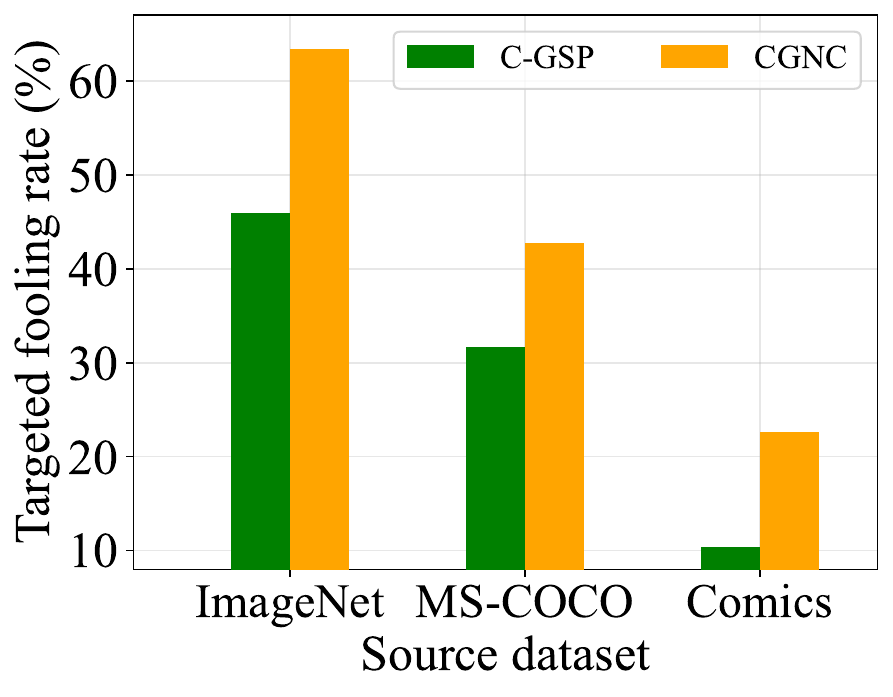}
  \centering
  \captionof{figure}{Fooling rates (from Res-152 to VGG-16) in attacking 8 target classes on cross-domain scenarios. We also provide the results on ImageNet as a comparison. }
\label{fig:cross}
\end{minipage}
\begin{minipage}[t]{.02\linewidth}
\quad
\end{minipage}
\begin{minipage}[t]{.48\linewidth}
  \centering
  \includegraphics[width=0.85\textwidth]{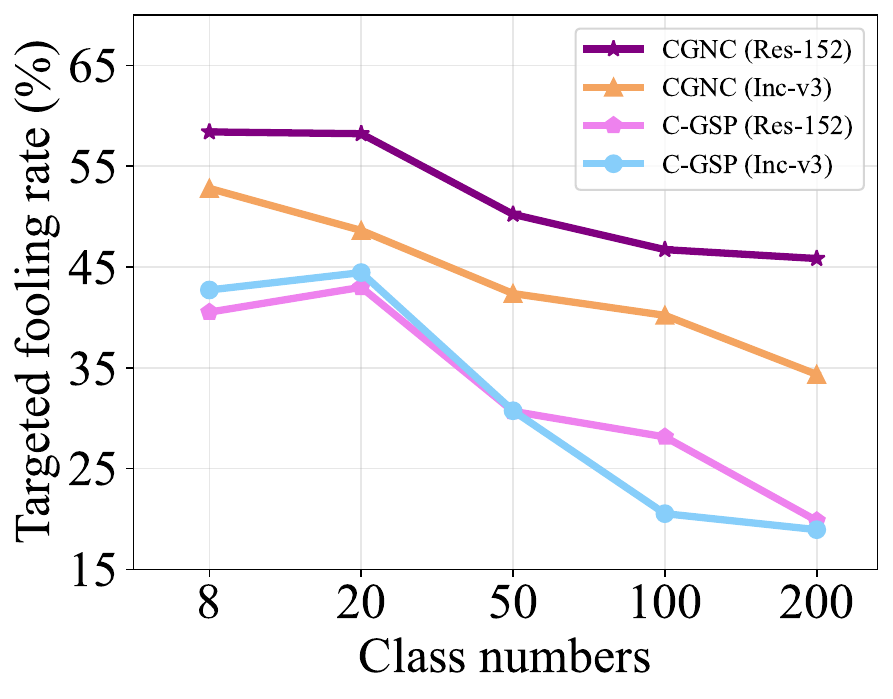}

  \captionof{figure}{Fooling rates of our CGNC with C-GSP on larger numbers of attacked classes regarding Res-152 and Inc-v3 as surrogate models. }
\label{fig:numbers}
\end{minipage}
\vspace{-1em}
 \end{figure*}

\begin{table}[t]
  \centering
  \caption{Ablation study of CGNC and its two variants on ImageNet NeurIPS validation set for 8 target classes. The substitute model is Res-152.}
    \resizebox{0.8\linewidth}{!}{\setlength{\tabcolsep}{3mm}{\begin{tabular}{cccccc}
    \toprule
    Method & VGG-16 & GoogleNet & Inc-v3 & Res-152 & DN-201 \\ \midrule
    CGNA-CA-t & 56.55 & 51.09 & 47.44 & 92.81 & 74.65 \\
    CGNA-CA & 56.64 & 54.29 & 49.73 & 93.34 & 75.99 \\
    CGNC  & 63.36 & 62.23 & 53.39 & 95.85 & 82.69 \\
    \bottomrule
    \end{tabular}%
  \label{tab:ablation}%
  }}
  \vspace{-1em}
\end{table}%

\subsection{Evaluation on More Scenarios}
\textbf{Cross-Domain scenarios.} Next, we explore the more realistic cross-domain scenarios \cite{naseer2019cross, naseer2021generating} where attackers do not know anything about the data distribution of the training set used by the black-box classifier. Based on this setting, attackers train the generator and generate adversarial perturbations using an auxiliary dataset that follows a different probability distribution from that of the target model. Specifically, we satisfy the cross-domain experimental setting using MS-COCO \cite{lin2014microsoft} and Comics \cite{comics} datasets respectively. MS-COCO is a large-scale image dataset widely used for object detection and semantic segmentation, while Comic is composed of a large number of comic images that can be regarded as a stylized version of ImageNet. 

To analyze the cross-domain transferability, we randomly select 1000 images from the source dataset to craft adversarial samples using the trained generator. Fig. \ref{fig:cross} shows the results of the cross-domain attacks transferring from three different source datasets to ImageNet. 
In general, the more challenging cross-domain datasets lead to varying degrees of reduction in the targeted fooling rate due to the domain gap, especially on the Comics dataset which differs greatly from the ImageNet distribution. 
Benefiting from the text guidance on the target category, CGNC still attains decent attack performance and remarkably outperforms C-GSP. This demonstrates that our method achieves better cross-domain transferability and is partly independent of the training dataset. 

In addition, since the SOTA single-target attacks \cite{naseer2021generating, feng2023dynamic, wang2023towards} require samples of the target class for loss computation, they are not applicable in the cross-domain scenarios where the source dataset lacks images of the target category. This represents an additional advantage of our method over single-target attacks.

\noindent\textbf{Larger Numbers of Target Classes.}
\label{sec:largerNums}
We then increase the number of target categories to verify the effectiveness of our method when handling plenty of classes. Specifically, we condition CGNC with more text inputs corresponding to the increased number of target categories. The performance is evaluated across the aforementioned six black-box models.

As mentioned before, single-target attacks become impractical for real-world classification systems with hundreds or thousands of target categories \cite{han2019once}. In comparison, our method effectively solves this issue and achieves great improvements in large numbers of target classes over C-GSP as in Fig. \ref{fig:numbers}, \eg, 14.71\% increase for 200 target classes proxy on the Res-152. 
This again reveals the significance of the fully exploited textual guidance from the CLIP model. 
Additionally, our method exhibits greater robustness to the varying numbers of target classes, \eg, the performance of our network on Inc-v3 demonstrates a smoother decrease compared to that of C-GSP. This further highlights the superiority of our method in reducing computation costs. For instance, when launching 1000-class targeted attacks using Inc-v3 as the surrogate model, our method requires training only 5 CGNCs with 200 conditions, yet it can achieve comparable performance to 20 generators with 50 conditions trained using C-GSP.

\subsection{Ablation Study}

We conduct ablation experiments on the ImageNet NeurIPS dataset to study the effect of the proposed techniques. Specifically, we introduce two variants of CGNC. CGNC-CA removes the cross-attention module from the original network, and CGNC-CA-t further modifies the conditioning mechanism by replacing the CLIP's text embedding with one-hot labels.

From the results in Table \ref{tab:ablation}, we can find that each aforementioned technique can further improve the attack success rates. Moreover, the remarkable improvement from CGNC-CA to CGNC also confirms that these cross-attention modules in the CA-Decoder can make better use of the text guidance provided by the CLIP model to enhance the targeted transferability of crafted perturbations.

 
\begin{table}[t]
\setlength{\tabcolsep}{8pt}
  \centering
  \caption{Comparison of our method with the SOTA single-target attacks. $^{\dagger}$ denotes the single-target variant of our CGNC obtained through the masked fine-tuning technique. $^{*}$ represents the white-box attacks.}
    \resizebox{0.95\linewidth}{!}{\begin{tabular}{c|c|c|c|c|c|c|c|c}
    \toprule
    Source   & Method   & Inc-v3   & Inc-v4   & Inc-Res-v2 & Res-152   & DN-121   & GoogleNet & VGG-16 \\
    \hline
    \multirow{5}[0]{*}{Inc-v3} & GAP      & 86.90$^{*}$    & 45.06    & 34.48    & 34.48    & 41.74    & 26.89    & 34.34 \\
             & CD-AP    & 94.20$^{*}$    & 57.60    & 60.10    & 37.10    & 41.60    & 32.30    & 41.70 \\
             & TTP      & 91.37$^{*}$    & 46.04    & 39.37    & 16.40    & 33.47    & 25.80    & 25.73 \\
             & DGTA-PI  & 94.63$^{*}$    & 67.95    & 55.03    & 50.50    & 47.38    & 47.67    & 48.11 \\
             & \cellcolor{black!15}CGNC$^{\dagger}$ & \cellcolor{black!10}\textbf{98.84}$^{*}$ & \cellcolor{black!10}\textbf{74.76} & \cellcolor{black!10}\textbf{64.48} & \cellcolor{black!10}\textbf{62.00} & \cellcolor{black!10}\textbf{78.94} & \cellcolor{black!10}\textbf{69.06} & \cellcolor{black!10}\textbf{70.74} \\
             \hline
    \multirow{5}[0]{*}{Res-152} & GAP      & 30.99    & 31.43    & 20.48    & 84.86$^{*}$    & 58.35    & 29.89    & 39.70 \\
             & CD-AP    & 33.30    & 43.70    & 42.70    & 96.60$^{*}$    & 53.80    & 36.60    & 34.10 \\
             & TTP      & 62.03    & 49.20    & 38.70    & 95.12$^{*}$    & 82.96    & 65.09    & 62.82 \\
             & DGTA-PI  & 66.83    & 53.62    & \textbf{47.61} & 96.48$^{*}$    & 86.61    & 68.29    & \textbf{69.58} \\
             & \cellcolor{black!10}CGNC$^{\dagger}$ & \cellcolor{black!10}\textbf{68.86} & \cellcolor{black!10}\textbf{69.45} & \cellcolor{black!10}45.71 & \cellcolor{black!10}\textbf{98.61}$^{*}$ & \cellcolor{black!10}\textbf{91.14} & \cellcolor{black!10}\textbf{69.83} & \cellcolor{black!10}68.05 \\
             \bottomrule
    \end{tabular}%
    }
  \label{tab:single}%
\end{table}%

\subsection{Comparison with Single-Target Attacks}
\label{sec:single_target}
Next, we compare the single-target variants of the proposed CGNC with multiple state-of-the-art single-target attacks. To obtain our single-target generators, we conduct the proposed masked fine-tuning to the trained conditional generator eight times using eight different text prompts of the target classes. 
Quantitative results in Table \ref{tab:single} indicate that our single-target enhanced generators outperform the competing single-target attacks in most cases. Particularly on the surrogate model of Inc-v3, CGNC$^{\dagger}$ achieves a notable increase of 15.36\% in the average black-box fooling rate compared to previous methods, 
demonstrating the significant effectiveness of the proposed masked fine-tuning. 
The MFT technique further enhances the single-target performance, thereby improving the adaptability and flexibility of our approach in scenarios with fewer target classes.

Note that these single-target methods require training 8 generators from scratch for 8 different target classes. In contrast, our method only needs to train a single multi-target generator and perform fine-tuning 8 times, each time with just a few epochs. Based on the typical experimental setup used in the SOTA single-target methods \cite{naseer2021generating, feng2023dynamic}, our strategy can diminish over 100 training epochs when targeting 8 classes, thus substantially mitigating the computational burden.
More experiments of single-target attacks against adversarially robust models and input preprocessing defenses are shown in the Appendix.

\section{Conclusion}
In this paper, we design a novel generative network CGNC, which improves multi-target transferable adversarial attacks by fully utilizing the rich prior within the CLIP as auxiliary semantic knowledge about the target category. To better incorporate the prior information, we introduce the cross-attention modules and efficiently condition the generator with CLIP's text embeddings. Through extensive experiments, we demonstrate the effectiveness of the proposed CGNC on multiple black-box target models in a variety of scenarios.
Moreover, we propose a masked fine-tuning technique to further enhance the CGNC in attacking a single class, which outperforms previous single-target methods in both efficiency and effectiveness. 
We hope that the proposed method can serve as a reliable tool for evaluating the model robustness under black-box setups, promoting further research on the vulnerability and robustness of DNNs.

\section*{Acknowledgments}
This work is supported in part by the National Natural Science Foundation of China under grant 62171248, 62301189, Guangdong Basic and Applied Basic Research Foundation under grant 2021A1515110066, the PCNL KEY project (PCL2021A07), and Shenzhen Science and Technology Program under Grant JCYJ20220818101012025, 
RCBS20221008093124061, GXWD20220811172936001.

\bibliographystyle{splncs04}
\bibliography{main}

\clearpage
\newpage

\appendix


\begin{figure*}[htbp]
\begin{center}
\includegraphics[width=1\linewidth]{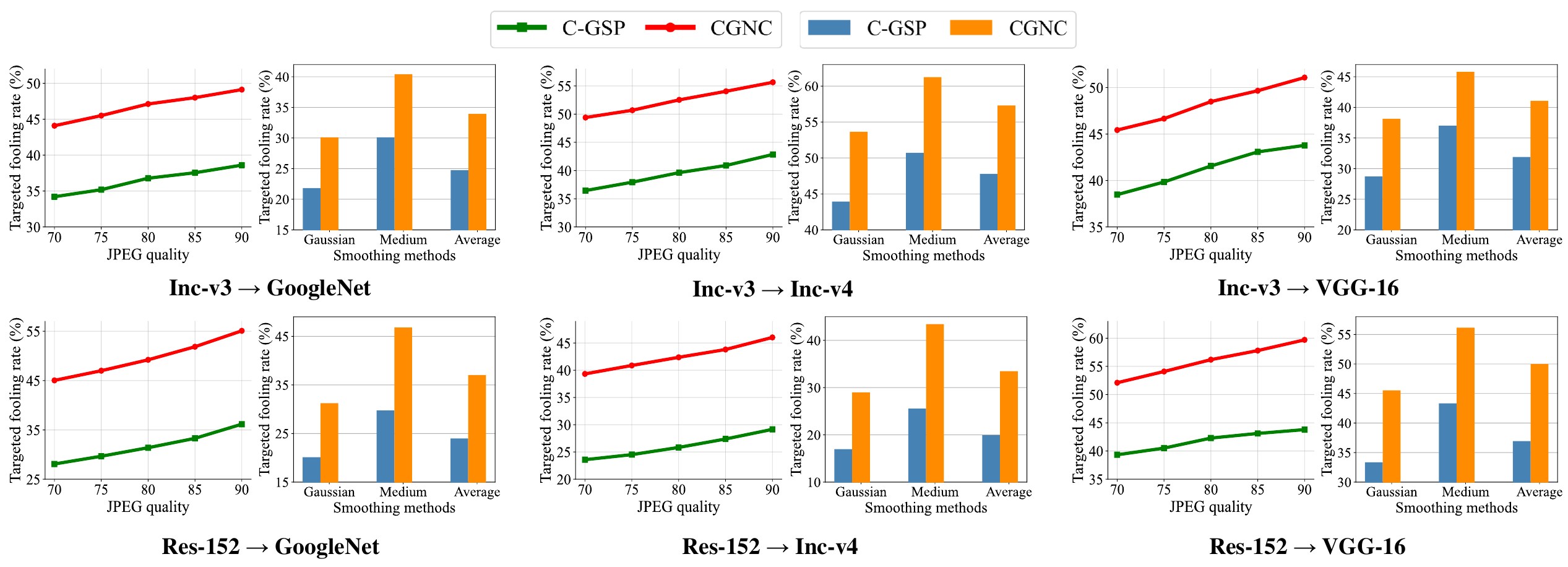}
\end{center}
\vspace{-1em}
\caption{Targeted transferability comparison of our CGNC and C-GSP \cite{yang2022boosting} on different victim models under various input processing defenses.}
\label{fig:multiinput}
\vspace{-3em}
\end{figure*}

\section{Additional Experiments}
We present additional experimental results to conduct a comprehensive comparison and in-depth analysis. Similarly, we follow \cite{yang2022boosting, feng2023dynamic} and adopt their used 8 classes as our target categories and compute the average attack success rates (ASR) on the 8 target classes as metrics. Unless stated otherwise, we use the ImageNet-NeurIPS (1k) dataset \cite{Nips2017} to evaluate the attack performance.

\subsection{Evaluation under Input Processing Defenses}
As mentioned before, we provide results on more victim models to compare our method and C-GSP \cite{yang2022boosting} under various input processing defenses.
Figure \ref{fig:multiinput} verifies that our CGNC consistently surpasses C-GSP under the considered defense strategies, revealing the effectiveness of our CLIP-empowered network.

\begin{table}[b]

\setlength{\tabcolsep}{8pt}
  \centering
  \caption{Attack success rates (\%) for multi-target attacks against regularly trained models on ImageNet validation set. * represents white-box attacks.}
    \resizebox{1\linewidth}{!}{\begin{tabular}{ccccccccc}
    \toprule 
    Source   & Method   & Inc-v3   & Inc-v4   & Inc-Res-v2    & Res-152  & DN-121   & GoogleNet       & VGG-16 \\ [0.05pt]
    \midrule
    \multirow{2}[0]{*}{Inc-v3} 
             & C-GSP    & 84.25$^*$    & 45.34    & 35.99    & 36.70    & 57.29    & 41.88    & 48.54 \\
             & CGNC& \textbf{96.59}$^*$& \textbf{57.82}&\textbf{46.84}&\textbf{44.13}&\textbf{65.90}&\textbf{53.40}&\textbf{56.27}\\
             \midrule
    \multirow{2}[0]{*}{Res-152} 
             & C-GSP    & 34.92    & 33.18    & 18.43    & 88.65$^*$    & 62.61    & 41.41    & 44.55 \\
             & CGNC&\textbf{56.00}&\textbf{50.37}&\textbf{32.26}&\textbf{96.44}$^*$&\textbf{86.69}&\textbf{63.84}&\textbf{63.90}\\
             \bottomrule
    \end{tabular}%
    }
  \label{tab:imagenetVal}%
\end{table}%

\subsection{Evaluation on ImageNet Validation Set}
For a more overall analysis, we compare our proposed CGNC and C-GSP \cite{yang2022boosting} on the whole ImageNet \cite{deng2009imagenet} validation set (50k samples). The experimental results are shown in Table \ref{tab:imagenetVal}. Evidently, our method stably achieves better transferability, with average improvements of 19.66{\%} and 9.77{\%} in black-box ASR using Res-152 and Inc-v3 as surrogate models respectively.

\begin{table}[t]
  \centering
  \caption{Comparison results on three black-box models under different perturbation budgets $\epsilon$. The surrogate model is Res-152.}
   \resizebox{0.95\linewidth}{!}{ \setlength{\tabcolsep}{2mm}{\begin{tabular}{cccccccccc}
    \toprule
    \multirow{2}[0]{*}{Method} & \multicolumn{3}{c}{VGG-16} & \multicolumn{3}{c}{Inc-v3} & \multicolumn{3}{c}{DN-121} \\ \cmidrule(lr){2-4} \cmidrule(lr){5-7} \cmidrule(lr){8-10}
          & $8/255$ & 12/255 & 16/255 & 8/255 & 12/255 & 16/255 & 8/255 & 12/255 & 16/255 \\ \midrule
    Logit & 2.71  & 5.91  & 9.20  & 1.65  & 4.70  & 10.10 & 2.86  & 6.62  & 12.70 \\
    SU    & 3.55  & 9.13  & 14.28 & 2.34  & 6.59  & 12.36 & 3.95  & 9.62  & 16.13 \\
    C-GSP & 15.48 & 32.11 & 45.90 & 10.43 & 23.98 & 37.70 & 31.66 & 56.79 & 64.20 \\
    Ours  & \textbf{21.46} & \textbf{46.28} & \textbf{63.36} & \textbf{15.04} & \textbf{37.35} & \textbf{53.39} & \textbf{45.83} & \textbf{73.05} & \textbf{85.66} \\ \bottomrule
    \end{tabular}%
    }
    }
  \label{tab:epsilon}%
\end{table}%

\subsection{Evaluation on Different Perturbation Budget $\epsilon$}
We then explore attacks under different $\epsilon$ values.  Specifically, we additionally consider smaller $\epsilon$ values of 8/255 and 12/255, where the adversarial perturbations are more imperceptible. The experimental results in Table \ref{tab:epsilon} reveal that our proposed network outperforms both the powerful iterative attacks Logit \cite{zhao2021success}, SU \cite{wei2023enhancing}, and the state-of-the-art (SOTA) multi-target generative attack C-GSP.

\subsection{Ablation analysis of CGNC}
In this section, we use Res-152 as the substitute model and present additional ablative experiments concerning our proposed CGNC to verify the contribution of each technique and investigate the influence of certain hyper-parameters.

\noindent\textbf{The effect of VL-Purifier.}
We first explore the influence of the VL-Purifier module. Specifically, we design CGNC-P that removes the VL-Purifier from the CGNC network. From Table \ref{tab:cgnc_ablation}, we find that directly incorporating CLIP's text embedding into the generator leads to serious performance degradation, which confirms the importance of this purifier module.

\begin{table}[t]
  \centering
  \caption{ASR of CGNC and its three variants. $^*$ denotes white-box attacks.}
    \resizebox{0.82\linewidth}{!}{\setlength{\tabcolsep}{2.8mm}{
    \begin{tabular}{cccccc} \toprule
    Architecture & VGG-16 & GoogleNet & Inc-v3 & Res-152 & DN-201 \\ \midrule
    CGNC  & 63.36 & 62.23 & 53.39 & 95.85$^{*}$ & 82.69 \\
    CGNC-P & 49.84   & 47.76   & 44.15   & 91.18$^{*}$   & 71.09 \\
    CGNC-F & 56.85 & 54.80  & 52.14 & 96.45$^{*}$ & 82.19 \\
    CGNC-t & 50.55 & 50.49 & 44.55 & 91.30$^{*}$  & 73.38 \\   \bottomrule
    \end{tabular}%
    }}
    \vspace{-1.5em}
  \label{tab:cgnc_ablation}%
\end{table}%

\begin{figure*}[t]
\begin{minipage}[t]{0.34\linewidth} 
  \centering
  \includegraphics[width=0.99\textwidth]{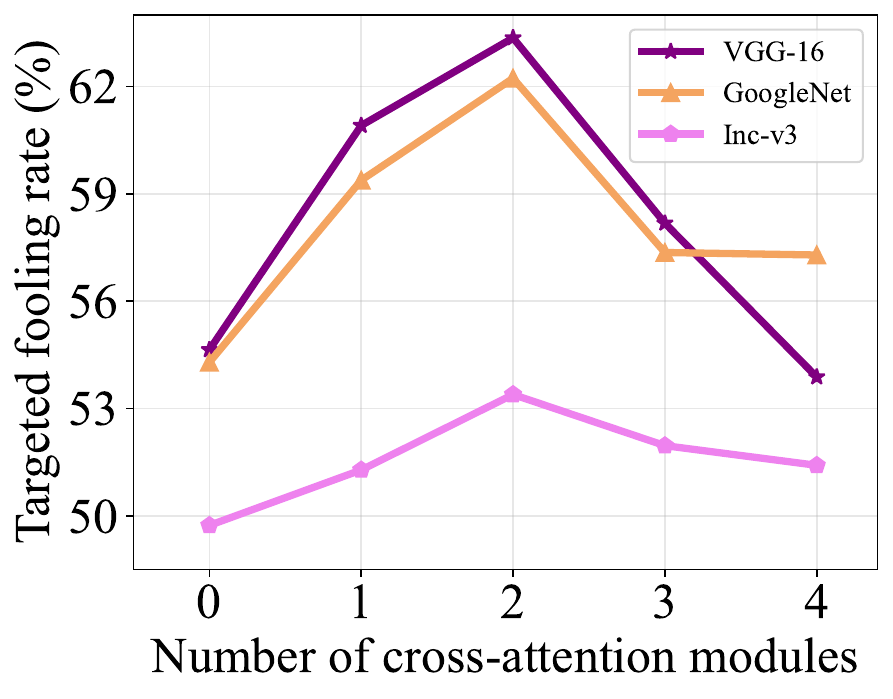}
  \centering
  \captionof{figure}{ASR on three target models with various numbers of cross-attention modules. }
  \label{fig:ca_modules}
\end{minipage}
\begin{minipage}[t]{.01\linewidth}
\quad
\end{minipage}
\begin{minipage}[t]{0.64\linewidth}
	\centering
	\begin{subfigure}{0.48\linewidth}
		\centering
		\includegraphics[width=\linewidth]{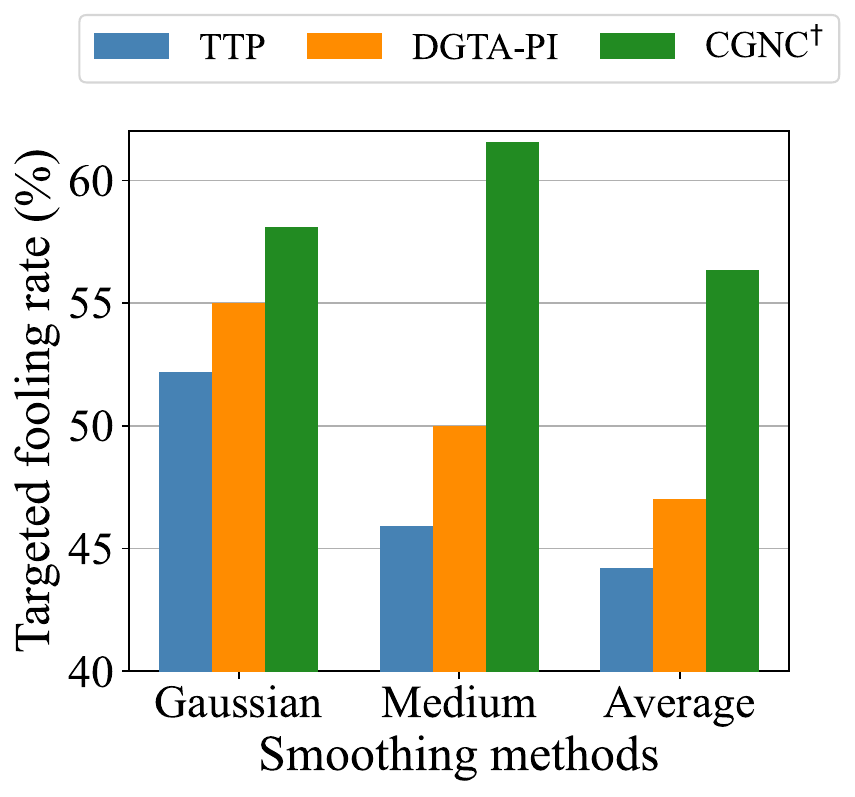}
		\caption*{}
	\end{subfigure}
	\centering
	\begin{subfigure}{0.505\linewidth}
		\centering
		\includegraphics[width=\linewidth]{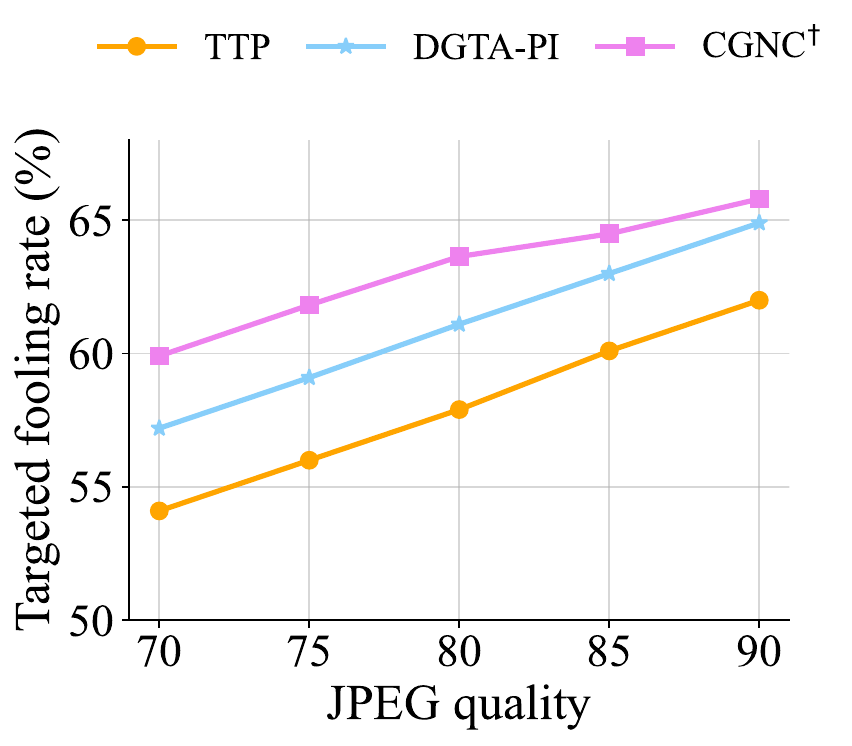}
		\caption*{}
	\end{subfigure}
	\centering
 \vspace{-1.2em}
	\caption{Comparison of our single-target variant CGNC$^{\dagger}$ with the SOTA single-target attacks under various input processing defenses. The victim model is VGG-16.}
 \label{fig:single_defense}
\end{minipage}
\vspace{-2em}
 \end{figure*}

\noindent\textbf{The effect of feature fusion.}
To verify the effectiveness of feature fusion operation in the F-Encoder, we introduce a variant CGNC-F which cancels the concatenate operation for feature fusion. The experimental results in Table \ref{tab:cgnc_ablation} validate the significance of the multi-modal feature fusion process.

\noindent\textbf{The effect of CLIP's text embedding.}
We analyze the effect of the text embedding by implementing a version CGNC-t that replaces all the text inputs with one-hot labels. The remarkable improvement from CGNC-t to CGNC shown in Table \ref{tab:cgnc_ablation} directly confirms the effectiveness of incorporating text information into the generator's architecture.

\noindent\textbf{Numbers of the cross-attention modules.}
We analyze the impact of the number of cross-attention modules on the attack performance. As illustrated in Fig. \ref{fig:ca_modules}, the generator exhibits optimal performance across all considered target models when employing two cross-attention modules. Consequently, we integrate two cross-attention modules into the backbone of the CA-Decoder.

\noindent\textbf{Scales of training data.} We adopt the same settings as previous generative attacks (\eg, CD-AP \cite{naseer2019cross}, TTP \cite{naseer2021generating}, C-GSP \cite{yang2022boosting}, and DGTA-PI \cite{feng2023dynamic}) 
and thus use the whole ImageNet training set to train generators. 
To investigate the influence of amount of training data, we further conduct training with different numbers of images. Tab. \ref{tab:data} shows that the scale of the training set indeed has a notable influence on the performance and our CGNC always outperforms C-GSP \cite{yang2022boosting}. 
\vspace{-1em}

\begin{table}[b]
\setlength{\tabcolsep}{10pt}
  \centering
  \vspace{-1.5em}
  \caption{ASR under various proportions of ImageNet training set.}
    \resizebox{0.8\linewidth}{!}{\begin{tabular}{ccccc}  \toprule
    Datset proportion  & 1/4 & 1/2 & 3/4 & 1   \\  \midrule
    C-GSP (Res-152) &   25.65    &  28.99     &  38.21     &  40.52   \\
    CGNC  (Res-152) &  \textbf{37.88}     &  \textbf{45.79}    & \textbf{51.20}    &  \textbf{58.40}   \\  \bottomrule
    \end{tabular}
    }
  \label{tab:data}%
\end{table}%

\subsection{More Comparison with Single-Target Attacks}
We provide more experimental results regarding Res-152 as the surrogate model to compare the single-target variant CGNC$^{\dagger}$ obtained through masked fine-tuning (MFT) with SOTA single-target methods, including GAP \cite{poursaeed2018generative}, CD-AP \cite{naseer2019cross}, TTP \cite{naseer2021generating}, and DGTA-PI\cite{feng2023dynamic}. 

\noindent\textbf{Comparison under Defense Strategies.} We consider the same defense strategies discussed in the main body of this manuscript. 
On attacking the adversarially robust model, our method achieves a notable average improvement of 4.37\% across six target models as shown in Table \ref{tab:single_robust}, demonstrating the excellent generalization ability of the proposed CGNC$^{\dagger}$.

For input defense strategies, Fig. \ref{fig:single_defense} shows that our CGNC$^{\dagger}$ also outperforms other methods when targeting models equipped with such defenses, especially for the input smoothing operations. It is also noteworthy that our method, which initially lags behind DGTA-PI \cite{feng2023dynamic} when attacking normally trained VGG-16, achieves a comprehensive lead after applying the smoothing operations and JPEG compression, highlighting the robustness and superiority of CGNC$^{\dagger}$ in handling various input-based defenses.

These results again indicate that although CGNC is designed for multi-target attacks, it can achieve better performance than these powerful single-target attack methods by simply fine-tuning it with a mask operation, revealing its great potential and scalability.

\noindent\textbf{Ablation analysis of the masked fine-tuning.}
To further verify the effectiveness of the proposed mask fine-tuning mechanism, we conduct ablation experiments and calculate the average ASR for each target class across the six black-box models. The results in Table \ref{tab:mask_ablation} illustrate the significance of both fine-tuning and patch-wise mask operation.

\begin{table}[t]
  \centering
  \caption{Comparison of the proposed CGNC$^{\dagger}$ with existing single-target attacks against target models with robust training mechanisms.}
  \vspace{-0.9em}
    \resizebox{0.85\linewidth}{!}{\setlength{\tabcolsep}{2.8mm}{\begin{tabular}{ccccccc} 
    \toprule
     Method & $\textrm{Inc-v3}_\textrm{adv}$ & $\textrm{IR-v2}_\textrm{ens}$ & $\textrm{Res50}_\textrm{SIN}$ & $\textrm{Res50}_\textrm{IN}$ & $\textrm{Res50}_\textrm{fine}$ & $\textrm{Res50}_\textrm{Aug}$ \\ \midrule
    GAP  & 5.72  & 4.51  & 7.33  & 71.04 & 83.64 & 52.07 \\
    CD-AP& 3.77  & 6.48  & 7.09  & 63.72 & 76.79 & 49.67 \\
    TTP  & 27.99 & 26.08 & 24.61 & 72.47 & 74.51 & 70.96 \\
    DGTA-PI & 31.10  & 30.07 & 27.70  & 77.13 & 80.55 & \textbf{76.78} \\
    CGNC$^{\dagger}$  & \textbf{31.55} & \textbf{33.63} & \textbf{33.31} & \textbf{88.34} & \textbf{89.74} & 72.96 \\    \bottomrule
    \end{tabular}%
    }}
  \label{tab:single_robust}%
\end{table}%

\begin{table}[t]
  \centering
  \caption{ASR of 8 different target classes. We compare the normal fine-tuning and our masked fine-tuning technique (\ie, CGNC$^{\dagger}$) for single-target attacks.}
    \vspace{-0.9em}
    \resizebox{0.89\linewidth}{!}{ \setlength{\tabcolsep}{2mm}{\begin{tabular}{cccccccccc} \toprule
    \multirow{2}[0]{*}{Source} & \multirow{2}[0]{*}{Method} & \multicolumn{8}{c}{Target class id} \\ \cmidrule{3-10}
          &       & 150   & 426   & 843   & 715   & 952   & 507   & 590   & 62 \\ \midrule
    \multirow{3}[0]{*}{Res-152} & CGNC  & 72.10 & 46.02 & 60.08 & 50.97 & 60.63 & 54.78 & 47.03 & 75.58 \\
          & Fine-tuning & 73.63 & 56.43 & 71.57 & 45.78 & 70.82 & 59.25 & 45.97 & 75.43 \\
          & MFT   & \textbf{78.38} & \textbf{63.32} & \textbf{76.12} & \textbf{56.47} & \textbf{78.40} & \textbf{64.18} & \textbf{49.65} & \textbf{84.20} \\ \midrule
    \multirow{3}[0]{*}{Inc-v3} & CGNC  & 64.27 & 46.17 & 47.78 & 38.82 & 60.32 & 52.65 & \textbf{51.05} & 61.63 \\
          & Fine-tuning & 70.23 & 61.72 & 72.43 & 48.30 & 64.43 & 68.12 & 42.65 & 56.03 \\
          & MFT   & \textbf{81.63} & \textbf{72.20} & \textbf{81.82} & \textbf{52.38} & \textbf{77.52} & \textbf{73.07} & 49.13 & \textbf{72.22} \\ \bottomrule
    \end{tabular}}}
    \vspace{-1.5em}
  \label{tab:mask_ablation}%
\end{table}%

\begin{table}[htbp]
  \centering
  \vspace{-1.5em}
    \setlength{\tabcolsep}{7pt}
  \caption{ASR on ViT-based models. The surrogate is Res-152.}
  \vspace{-0.9em}
    \resizebox{0.95\linewidth}{!}{\begin{tabular}{ccccccc} \toprule
    Method & ViT-B/16 & CaiT-S/24 & Visformer-S & DeiT-B & LeViT-256 & TNT-S \\ \hline
    C-GSP \cite{yang2022boosting} & 11.78 & 32.00    & 36.60  & 35.58 & 37.85 & 31.00    \\
    CGNC  & \textbf{19.46} & \textbf{54.56} & \textbf{58.70}  & \textbf{59.90}  & \textbf{57.53} & \textbf{48.40} \\   \bottomrule
  \end{tabular}%
  \label{tab:ViT}%
  }
  \vspace{-3em}
\end{table}%

\subsection{Attacks on Transformer-based models.}
We also evaluate on six ViT-based models in Tab. \ref{tab:ViT}. The results reveal that our CGNC also consistently exhibits better performance than C-GSP \cite{yang2022boosting} on Transformer-based models. 
\vspace{-1em}

\section{Limitations \& Future work.}

In this paper, we adopt a simple yet effective text template "a photo of a \{class\}" recommended by CLIP \cite{radford2021learning}, which has been proven effective in a variety of tasks. However, due to the excessive reliance \cite{huang2023sentence} on the statistical features of 'photo', this text template may limit the transferability performance to a certain extent, particularly for target datasets with stylized images.
Future research can consider introducing more accurate or detailed text as the description of the target class, such as the recommended list of eighty templates of text prompt by CLIP \cite{radford2021learning}, \eg, "a sculpture of a \{\}", "an art of \{\}". They can use some of their averaged representations as the generic representations of the target class.  Another promising approach is to choose a related pre-training task (\eg, classification) and use prompt learning \cite{zhou2022learning} to acquire the representation of the target category. These learned prompts can better represent the target class and distinguish features from different categories.

\section{More Visualization}
\begin{figure*}[htbp]
\begin{center}
\includegraphics[width=1\linewidth]{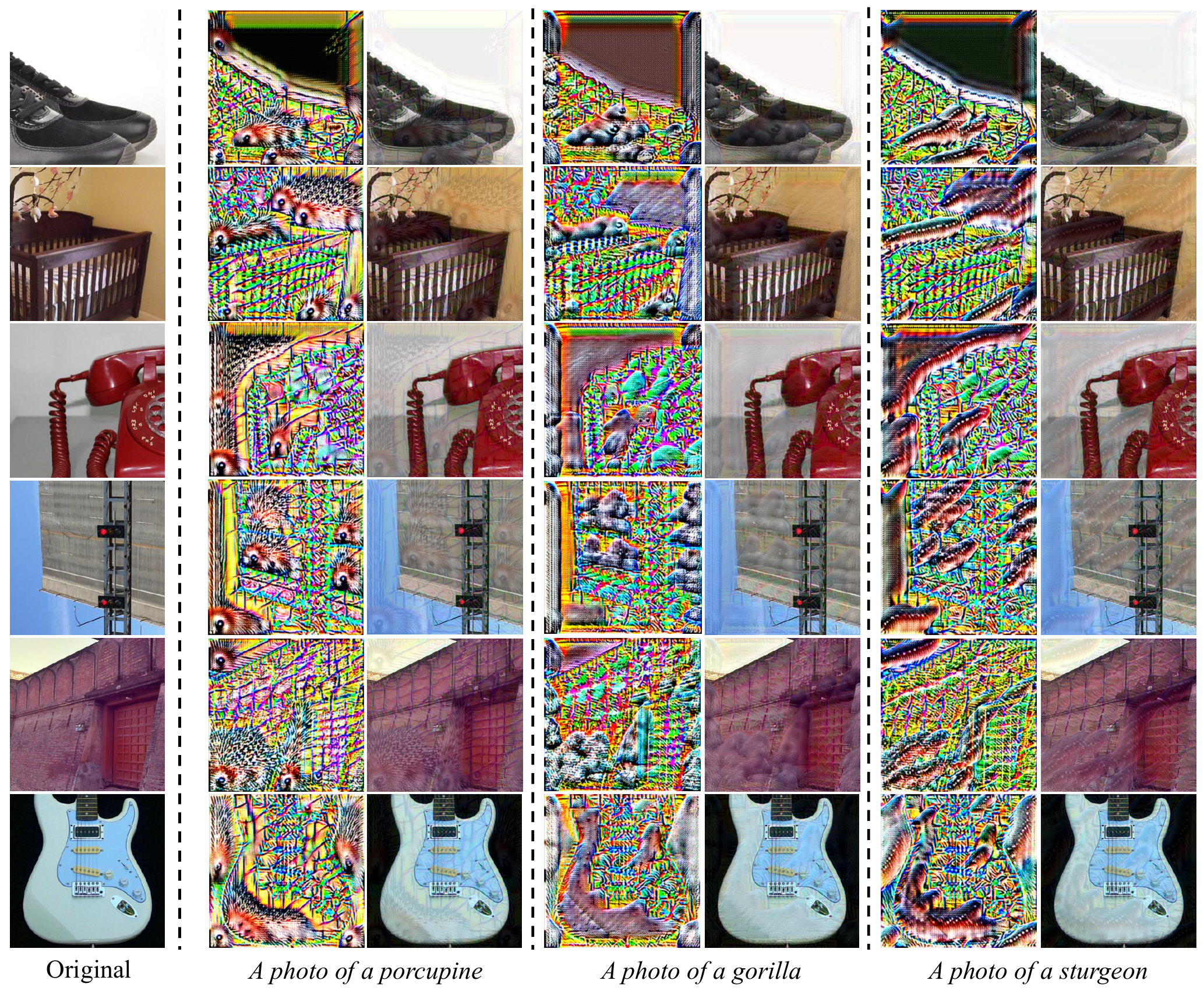}
\end{center}
\vspace{-1em}
\caption{Visualization of the generated perturbations and adversarial samples.}
\label{fig:visulization}
\vspace{-3em}
\end{figure*}

We provide more visualization results of generated perturbations and adversarial samples in Fig. \ref{fig:visulization}. The generated perturbations carry rich semantic patterns of the target class, and as we change the input text condition, the generated patterns vary accordingly to the target class. This once again demonstrates the effectiveness of our method in modeling the target features, as well as the success of conditioning the generator with CLIP's text embeddings.

\end{document}